\documentclass[utf8]{frontiersSCNS} % for Science, Engineering and Humanities and Social Sciences articles
\usepackage{url,hyperref,lineno,microtype,subcaption}
\usepackage[onehalfspacing]{setspace}
\usepackage{amsmath,amssymb}
\usepackage{algorithm}
\usepackage{algorithmic}
\usepackage{bm}

\newlength\myindent%インデントコマンド
\makeatletter
  \newcommand{\figcaption}[1]{\def\@captype{figure}\caption{#1}}
  \newcommand{\tblcaption}[1]{\def\@captype{table}\caption{#1}}
\makeatother

\def\firstAuthorLast{Hagiwara {et~al.}} %use et al only if is more than 1 author
\def\Authors{Yoshinobu Hagiwara\,$^{1,*}$, Hiroyoshi Kobayashi\,$^{1}$, Akira Taniguchi\,$^{1}$, and Tadahiro Taniguchi\,$^{1}$}
\def\Address{$^{1}$Emergent Systems Laboratory, College of Information Science and Engineering, Ritsumeikan University, Shiga, Japan}
\def\corrAuthor{Yoshinobu Hagiwara}
\def\corrEmail{yhagiwara@em.ci.ritsumei.ac.jp}

\begin{document}
\onecolumn
\firstpage{1}

\title[Symbol Emergence as an Interpersonal Multimodal Categorization]{Symbol Emergence as an Interpersonal Multimodal Categorization} 

\author[\firstAuthorLast ]{\Authors} %This field will be automatically populated
\address{} %This field will be automatically populated
\correspondance{} %This field will be automatically populated

\extraAuth{}% If there are more than 1 corresponding author, comment this line and uncomment the next one.
%\extraAuth{corresponding Author2 \\ Laboratory X2, Institute X2, Department X2, Organization X2, Street X2, City X2 , State XX2 (only USA, Canada and Australia), Zip Code2, X2 Country X2, email2@uni2.edu}

\maketitle

\begin{abstract}
This study focuses on category formation for individual agents and the dynamics of symbol emergence in a multi-agent system through semiotic communication.
Semiotic communication is defined, in this study, as the generation and interpretation of signs associated with the categories formed through the agent's own sensory experience or by exchange of signs  with other agents. From the viewpoint of language evolution and symbol emergence, organization of a symbol system in a multi-agent system (i.e., agent society) is considered as a bottom-up and dynamic process, where individual agents share the meaning of signs and categorize sensory experience. A constructive computational model can explain the mutual dependency of the two processes and has mathematical support that guarantees a symbol system’s emergence and sharing within the multi-agent system. In this paper, we describe a new computational model that represents symbol emergence in a two-agent system based on a probabilistic generative model for multimodal categorization. It models semiotic communication via a probabilistic rejection based on the receiver's own belief. 
We have found that the dynamics by which cognitively independent agents create a symbol system through their semiotic communication can be regarded as the inference process of a hidden variable in an interpersonal multimodal categorizer, i.e., the complete system can be regarded as a single agent performing multimodal categorization using the sensors of all agents, if we define the rejection probability based on the Metropolis-Hastings algorithm.
The validity of the proposed model and algorithm for symbol emergence, i.e., forming and sharing signs and categories, is also verified in an experiment with two agents observing daily objects in the real-world environment. 
In the experiment, we compared three communication algorithms: no communication, no rejection, and the proposed algorithm.
The experimental results demonstrate that our model reproduces the phenomena of symbol emergence, which does not require a teacher who would know a pre-existing symbol system. Instead, the multi-agent system can form and use a symbol system without having pre-existing categories.

% a primary goal, the abstract should render the general significance and conceptual advance of the work clearly accessible to a broad readership. References should not be cited in the abstract. Leave the Abstract empty if your article does not require one, please see \href{http://www.frontiersin.org/about/AuthorGuidelines#SummaryTable}{Summary Table} for details according to article type. 

%\tiny
% \keyFont{ \section{Keywords:} keyword, keyword, keyword, keyword, keyword, keyword, keyword, keyword} %All article types: you may provide up to 8 keywords; at least 5 are mandatory.
\end{abstract}

\section{Introduction}
Language plays a crucial role in sharing information between people by using semiotic communication  (i.e., exchanging signs).
However, it is still a mysterious problem in language evolution: how the symbol system of the language has  emerged through semiotic communication.
The semiotic communication in this study is defined as the generation and interpretation of signs associated with the categories formed through the agent's own sensory experience or by exchanging signs  with other agents. Although sharing and forming the symbols representing the sensory experience is only a part of the function of language, obtaining a computational explanation of emergence of a symbol system in the real-world environment (i.e., only from individual sensory-motor experience and exchanging signs) is important and challenging. 

A fundamental study on language evolution~\citet{Chomsky75} focused on thoughts and concept formation within individuals, and advocated the concept of generative grammar  that explains the syntactic ability required for languages.
\citet{Tomasello99} focused on human cooperative communication  and proposed a model to determine the communication ability for sharing intentions with others.
As a hypothesis that handles concept formation and intention sharing in an integrated manner, \citet{Okanoya07} advocated a mutual segmentation hypothesis of sound strings and situations based on co-creative communication.
Our study was conducted by a constructive approach inspired by the Okanoya's hypothesis, and it deals with the concept formation in individuals and intention sharing in multi-agent systems simultaneously.
Constructive approaches are highly advantageous to understanding complex systems and are also useful for studying evolutionary linguistics~\citep{Steels97, Hashimoto99}. \citet{mLDA,TANIGUCHI2018166}  developed integrative probabilistic generative models for multimodal categorization and word discovery by a robot (using multimodal sensorimotor information obtained by a robot, and speech signals). However, their model only explains individual learning, and they implicitly presume that a teacher has a fixed and static symbol system. The goal of this paper is to provide a computational model that not only categorizes sensory information but also shares the meaning of signs within the multi-agent system. This model provides a clear view of symbol emergence as an interpersonal multimodal categorization. 

In studies on language emergence in multi-agent systems,
\citet{Kirby99} showed that the language exchanged between agents involving repeated generation alternation is gradually structured  in a simulation model.
\citet{Morita12} showed in simulation experiments that semiotic communication systems emerge from interactions that solve collaborative tasks.
\citet{Lazaridou16} proposed a framework for language learning that relies on multi-agent communication for developing interactive machines (e.g., conversational agents).
\citet{Lee17} proposed a communication game in which two agents, native speakers of their own respective languages, jointly learn to solve a visual referential task. 
\citet{Graesser19} proposed a computational framework in which agents equipped with communication capabilities simultaneously play a series of referential games, where agents are trained by deep reinforcement learning.
These studies achieved language emergence in multi-agent systems by using a computational model.
However, interaction with the real-world environment through one's own sensory information without pre-existing categories (i.e., internal representations) was not discussed in these studies.

In studies on language evolution based on a constructive approach using robots, 
Steels performed experiments on a self-organizing spatial vocabulary~\citep{Steels95} and perceptually grounded categories through language for color ~\citep{Steels05}. Their research considered how people talk about the location of objects and places~\citep{Steels08}. The series of his studies using robots are summarized as the talking heads experiment~\citep{Steels15}.

In studies beyond the talking heads experiment, 
\citet{AIBO} performed experiments with robots in ``AIBO's first words: the social learning of language and meaning''. 
\citet{De06} proposed a cross-situational learning algorithm for damping homonymy in the guessing game. 
Spranger proposed a perceptual system for language game experiments~\citep{Spranger12} and performed the evolution of grounded spatial language~\citep{Spranger11,Spranger14}. \citet{Bleys15} proposed language strategies for the domain of color. 
\citet{Matuszek18} proposed grounded language learning, where robotics and natural language processing meet.
These studies focused on the symbol grounding in the language game and built the foundation of constructive studies on language evolution.
However, the modeling of bottom-up internal representation learning from sensory-motor information within individuals has not been discussed.

\citet{Tadahiro16} introduced a concept of {\it symbol emergence system}, which is a multi-agent system that dynamically organizes a symbol system, for example , by physical interaction with the environment and semiotic communication with other agents. A symbol emergence system can be regarded as an emergent system: a complex system that has an emergent property.
Internal representations (e.g. object categories) are formed in individual agents' cognitive systems under the influence of a symbol system shared among a multi-agent system. A symbol system shared in a society is organized under the influence of internal representations formed by individual agents. The interdependency is crucial for symbol emergence in a multi-agent system. However, so far, there is a lack of a computational model that would describe the mutual dependency and would be evaluated in experiments in the real-world environment.  

In studies of bottom-up concept formation and word grounding based on the sensory-motor information of a robot, \citet{Nakamura09} proposed a model for grounding word meanings in multimodal concepts, and \citet{Ando13} proposed a model of hierarchical object concept formation based on multimodal information.
Several methods that enable robots to acquire words and spatial concepts based on sensory-motor information in an unsupervised manner have been proposed~\citep{Isobe17,Akira17}. Hagiwara proposed a Bayesian generative model to acquire the hierarchical structure of spatial concepts based on the sensory-motor information of a robot~\citep{Hagiwara16,Hagiwara18}. 
These studies focused on language acquisition by a robot from a person who gives speech signals to the robot, and enables robots to discover words and categories based on their embodied meanings from raw sensory-motor information (e.g., visual, haptic, auditory, and acoustic speech information). They presume that a person has knowledge about categories and signs representing the categories, i.e., a symbol system shared in the society. Therefore, these computational models cannot 
be considered as a constructive model of symbol emergence systems. 
These studies have not dealt with the dynamics of emerging symbols while agents form categories based on sensory-motor information .

\begin{figure}[bt!]
	\begin{center}
	\includegraphics[width=\linewidth]{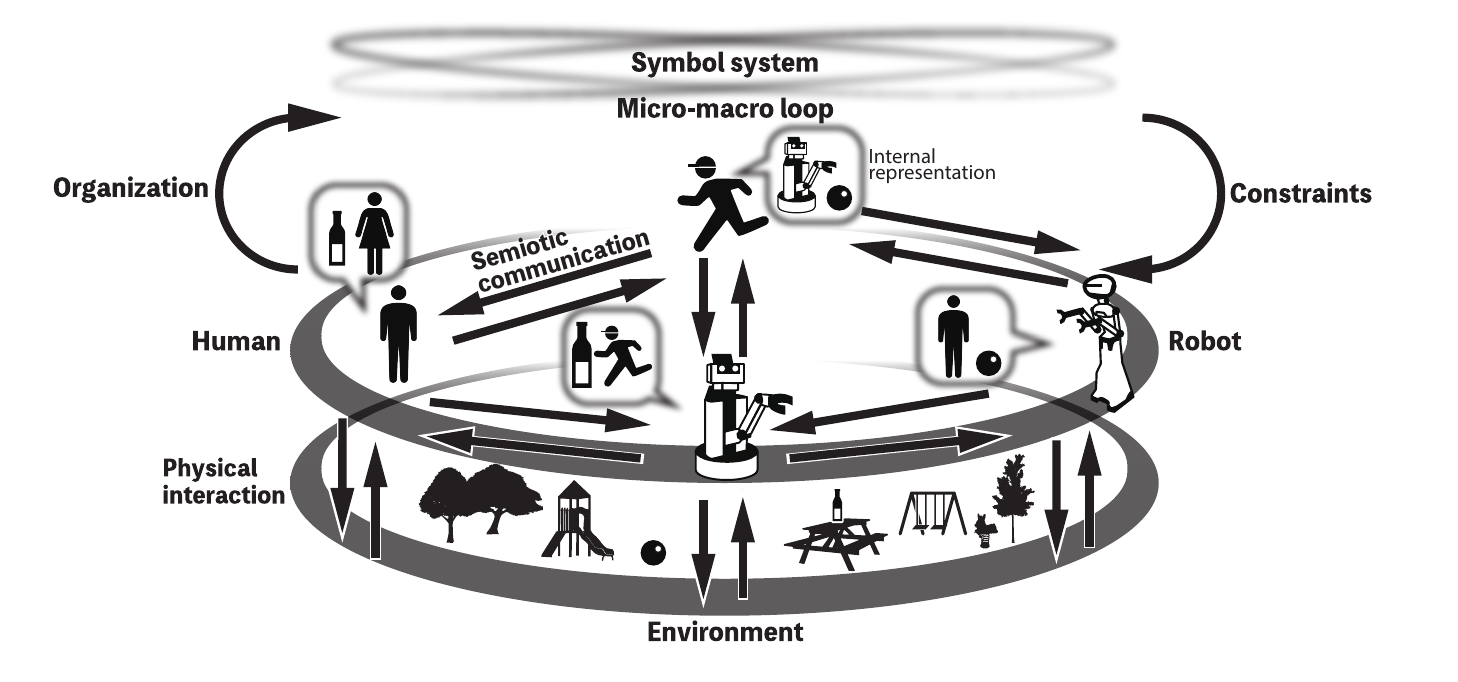}
	\caption{Overview of the symbol emergence systems~\cite{Tadahiro16}}
	\label{fig:ser}
	\end{center}
\end{figure}

A concept of symbol emergence in cognitive systems is surveyed in \citep{TadahiroTaniguchi2018}. A symbol emergence system is socially self-organized through both semiotic communications with autonomous cognitive developmental agents and physical interactions with the environment, as shown in Figure~\ref{fig:ser}. The figure represents a symbol emergence system.
Note that a symbol system cannot be controlled by anyone, but all individuals are constrained by an emergent and shared symbol system. In addition, all of them contribute to creating the socially shared symbol system. To understand this phenomena, the coupled dynamics of both a symbol system shared between the agents and the internal representation systems of individuals has to be modeled with a constructive and computational approach.
Computational models for category formation and lexical acquisition by a robot have been studied and proposed.
Studies on language evolution and symbol emergence have different challenges, in addition to concept formation and lexical acquisition. They have to deal with organization of a symbol system in the society itself (i.e., the symbol system is unstable), in contrast with studies on category formation and lexical acquisition, where the system can be considered as static and stable.

This study focuses on the symbol emergence in a multi-agent system and the category formation in individual agents through semiotic communication, which is the generation and interpretation of symbols associated with the categories formed using the agent's sensory information. 
The main contributions of this paper are as follows.
\begin{itemize}
 \item We propose a constructive computational model that represents the dynamics of a symbol emergence system by using probabilistic models for multimodal categorization and message passing based on the Metropolis-Hastings (M-H) algorithm. The model represents mutual dependency of individual categorization and formation of a symbol system in a multi-agent system. 
 \item We show that our model representing a multi-agent system and symbol emergence among agents can be regarded as a single agent and a single multimodal categorizer, i.e., an interpersonal categorizer, mathematically. We prove that the symbol emergence in the model is guaranteed to converge. 
 \item We evaluate the proposed model of the symbol emergence and category formation from an experiment by using two agents that can obtain visual information and exchange signs in the real-world environment. The results show the validity of our proposed model.
\end{itemize}

The rest of this paper is structured as follows. Section 2 describes the proposed model and inference algorithm for representing the dynamics of symbol emergence and category formation in multi-agent systems. Section 3 presents experimental results, verifying the validity of the proposed model and inference algorithm on the object categorization and symbol emergence. Finally, Section 4 presents conclusions.

\section{Proposed model and inference algorithm}
\subsection{Expansion of a multimodal categorizer from personal to interpersonal}

The computational model that we propose in this paper is based on a key finding that a probabilistic generative model of multimodal categorization can be divided into several sub-modules of probabilistic generative models for categorization  and message passing between the sub-modules. This idea of dividing a large probabilistic generative model for developing cognitive agents and re-integrating them was firstly introduced as a SERKET framework~\citep{Serket}. However, their idea was only applied to creating a single agent. We found that the idea can be used for modeling multi-agent systems and is very suitable for modeling dynamics of a symbol emergence system. 

\begin{figure}[bt!]
	\begin{center}
	\includegraphics[width=\linewidth]{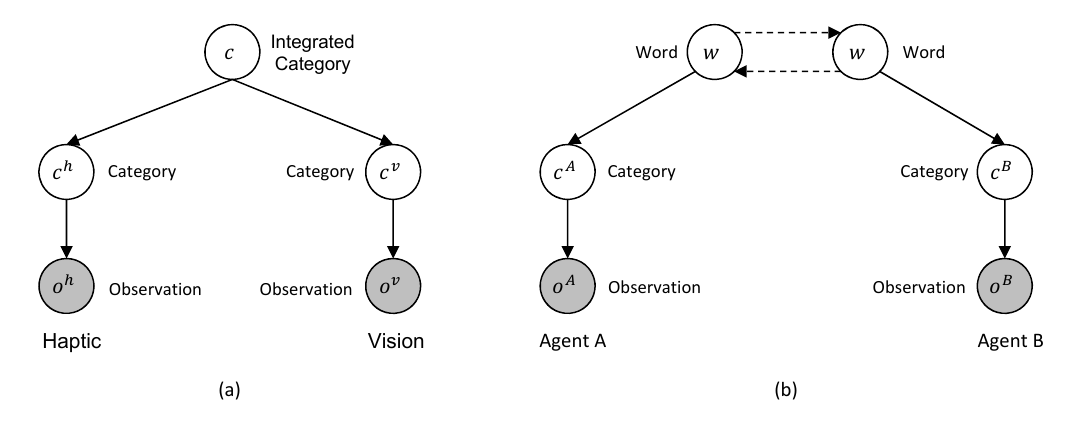}
	\caption{The expansion of a multimodal categorizer from personal to interpersonal: (a) shows a generative model of a personal multimodal categorizer between haptics and vision, and (b) shows a generative model of an inter-personal multimodal categorizer between the agents. Dashed lines in (b) show communication between agents. The parameters of these models are simplified.}
	\label{fig:categorizer}
	\end{center}
\end{figure}

We modeled the symbol emergence in a multi-agent system and the category formation in individual agents as a generative model by expanding a personal multimodal categorizer (see Figure~\ref{fig:categorizer} (a)) to an interpersonal multimodal categorizer  (see Figure~\ref{fig:categorizer} (b)).
First, (a) shows a personal multimodal categorizer, which is a generative model with an integrated category $c$ as a latent variable and sensor information from haptics and vision as observations $o^h$ and $o^v$. The model is a simple version of multimodal latent Dirichlet allocation  used as an object categorizer in the previous studies ~\citep{Nakamura09,Ando13}. Next, (b) shows an interpersonal multimodal categorizer in which two agents are modeled as a collective intelligence, with word $w$ as a latent variable, and sensor information from agent A and B as observations $o^A$ and $o^B$. As shown in Figure~\ref{fig:categorizer}, the model generating observations through categories on each sensor from an integrated concept in an agent can be extended as the model generating observations through categories on each agent from a word in a multi-agent system.

Figure~\ref{fig:categorizer} (a) represents a graphical model for probabilistic generative model  multimodal categorization, e.g., \citet{mLDA}. It can integrate multimodal information, e.g., haptics and visual information, and form categories. Index of category is represented by $c$ in this figure. 
Following the SERKET framework~\citep{Serket}, we can divide the model into two modules and a communication protocol for a shared node. Here, $c$ is shared by the two modules and the node is renamed by $w$. We regard an index $w$ as an index of word. In this case, if we regard the two separated modules as two individual agents (i.e., agent A and agent B), the communication between the two nodes can be considered as exchange of signs (i.e., words). As we see later, we found that, if we employ the Metropolis-Hastings algorithm, which is one of the communication protocols that the original SERKET paper proposed, the communication protocol between the nodes can be considered as semiotic communication between two agents. Roughly speaking, the communication is described as follows. Agent A recognizes an object and generates words for Agent B. If the word is consistent to the belief of Agent B, then Agent B accepts the naming with a certain probability; otherwise, Agent B rejects the information, i.e., does not believe the meaning. If the rejection and acceptance probability of the communication is the same as the probability of the M-H algorithm, the posterior distribution over $w$, i.e., symbol emergence among the agents, is theoretically the same as  the posterior distribution over $c$, i.e., interpersonal categorization. 

\subsection{Generative process on the interpersonal multimodal categorizer}

\renewcommand{\arraystretch}{1.3}
\begin{figure}[bt!]
\begin{tabular}{cc}
  \begin{minipage}{.50\textwidth}
	\centering
	\includegraphics[width=\linewidth]{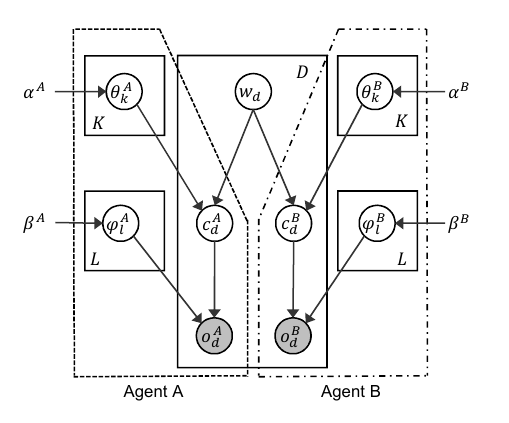}
    \figcaption{Graphical model of the proposed interpersonal multimodal categorizer}
    \label{fig:model}
  \end{minipage}
  \begin{minipage}{.50\textwidth}
    \begin{center}
    \tblcaption{Definition of variables in the proposed interpersonal multimodal categorizer}
    \label{tb:model}
    \renewcommand{\arraystretch}{1.7}
    \begin{tabular}{|c|l|} \hline
      $w_d$ & Index of the word  \\ \hline
      $c_d^A,c_d^B$ & Index of the category  \\ \hline
      $o_d^A,o_d^B$ & Observations of agents A and B \\ \hline
      $\phi_l^A,\phi_l^B$ & Parameters of the multinomial distribution \\ \hline
      $\theta_k^A,\theta_k^A$ & Parameters of the multinomial distribution\\ \hline
      $\alpha,\beta$ & Hyperparameters for $\theta,\phi$ \\ \hline
      $K$ & Number of words  \\ \hline
      $L$ & Number of categories  \\ \hline
      $D$ & Number of data points  \\ \hline
    \end{tabular}
    \end{center}
  \end{minipage}
\end{tabular}
\end{figure}
This subsection describes the generative process of the interpersonal multimodal categorizer.
Figure~\ref{fig:model} shows the graphical model is a single graphical model. However, following the SERKET framework (see Figure~\ref{fig:categorizer}), it can be owned by two different agents separately. The right and left parts indicated with a dashed line in Figure~\ref{fig:model} show the parts owned by agents A and B, respectively.
Figure~\ref{fig:model} and Table~\ref{tb:model} show the graphical model and the parameters of a proposed interpersonal multimodal categorizer, respectively.

Index $w_d$ (of word $w$)  connects the agents A and B as a hidden variable to generate the index of a category $c_d$ from the parameter of multinomial distribution $\theta_k$ in each agent.
$o_d^A$ and $o_d^B$ are observations on data point $d$ obtained from the sensors attached to the agents A and B, respectively.
$c_d^A$ and $c_d^B$ are indices of a category allocated to an observation $o_d^A$ and $o_d^B$, respectively.
$\phi_l^A$ and $\phi_l^B$ are the parameters of multinomial distributions to generate observations $o_d^A$ and $o_d^B$ based on categories $c_d^A$ and $c_d^B$.
$\alpha$ and $\beta$ are the hyperparameters of $\theta$ and $\phi$.
$K$ is the number of words in the word dictionary that a robot has.
$L$ is the number of categories.
$D$ is the number of observed data points.
The multinomial distribution is denoted as ${\rm Multi}(\cdot )$, and
the Dirichlet distribution is denoted as ${\rm Dir}(\cdot )$.

The generative process of the interpersonal multimodal categorizer is described as follows.

The parameters $\phi_l^A$ and $\phi_l^B$ of multinomial distributions on each category $(l \in L)$ are shown as follows: % in the table
\begin{equation}
    \phi_l^A \sim {\rm Dir}(\beta^A),
\end{equation}
\begin{equation}
    \phi_l^B \sim {\rm Dir}(\beta^B).
\end{equation}

The parameters $\theta_k^A$ and $\theta_k^B$ of multinomial distributions on each word $(k \in K)$ are shown as follows: % in the table
\begin{equation}
    \theta_k^A \sim {\rm Dir}(\alpha^A),
\end{equation}
\begin{equation}
    \theta_k^B \sim {\rm Dir}(\alpha^B).
\end{equation}

The following operations from (\ref{eq:sample_oda}) to (\ref{eq:sample_cdb}) are repeated for each data point $(d \in 1,2,...,D)$: 
\begin{itemize}
    \item Observations $o_d^A$ and $o_d^B$ generated from categories $c_d^A$ and $c_d^B$ are shown as follows:
    \begin{equation}
        o_d^A \sim {\rm Multi}(\phi_{c_d^A}^A),
        \label{eq:sample_oda}
    \end{equation}
    \begin{equation}
        o_d^B \sim {\rm Multi}(\phi_{c_d^B}^B).
        \label{eq:sample_odb}
    \end{equation}    
    \item Indices of categories $c_d^A$ and $c_d^B$ generated from word $w_{d}$ are shown as follows:
    \begin{equation}
        c_d^A \sim {\rm Multi}(\theta_{w_d}^A),
        \label{eq:sample_cda}
    \end{equation}
    \begin{equation}
        c_d^B \sim {\rm Multi}(\theta_{w_d}^B).
        \label{eq:sample_cdb}
    \end{equation} 
\end{itemize}

Theoretically, the generative model is a type of pre-existing model for multimodal categorization~\citep{Serket,mLDA} for an individual agent. In this paper, we assume that the graphical model is  representing the symbol emergence in a multi-agent system.

\subsection{Communication protocol as an inference algorithm on the interpersonal multimodal categorizer} 
This subsection describes the protocol of semiotic computation between two agents and cognitive dynamics of categorization in individual agents. As a whole, the total process can be regarded as a model of symbol emergence in the multi-agent system. Additionally, the total process can be regarded as an inference process of the  probabilistic generative model integrating the two agents' cognitive systems (Figure~\ref{fig:model}).

\subsubsection{Gibbs sampling}
First, to introduce our proposed model, we describe an ordinary Gibbs sampling algorithm for the integrative probabilistic generative model in Figure~\ref{fig:model}. Gibbs sampling algorithm is widely used for multimodal categorization and language learning in robotics. 
Gibbs sampling~\citep{Gibbs} is known as a type of Markov chain Monte Carlo (MCMC) algorithm for inferring latent variables in probabilistic generative models.

Algorithm~\ref{alg1} shows the inference algorithm on the model of Figure~\ref{fig:model} using Gibbs sampling. In the algorithm~\ref{alg1}, $i$ shows the number of iterations; ${\bf O}^A$ and ${\bf O}^B$ denote a set of all observations in agents A and B, respectively; ${\bf C}^A$ and ${\bf C}^B$ denote a set of all categories in agents A and B, respectively; and $W$ denotes a set of all words.
In line 14 of Algorithm~\ref{alg1}, word $w_d^{[i]}$ is sampled from the product of probability distributions $P(c_d^{A[i]} \mid \theta_k^{A[i]})$ and $P(c_d^{B[i]} \mid \theta_k^{B[i]})$ based on parameters $\theta_k^{A[i]}$ and $\theta_k^{B[i]}$ in agents A and B.

If an agent can observe both $\theta_k^{A[i]}$ and $\theta_k^{B[i]}$, which are internal representations of each agent, this algorithm can work. However,  Agent A cannot observe  $\theta_k^{B[i]}$, or Agent B cannot observe  $\theta_k^{A[i]}$. Therefore, no agent can perform Gibbs sampling in this multi-agent system. In this sense, this is not a valid cognitive model for representing the symbol emergence between two agents.
\begin{algorithm}
\caption{Gibbs sampling algorithm}
\label{alg1}
    \begin{algorithmic}[1]
    \STATE Initialize all parameters
    \FOR{$i = 1$ to $I$}
    %\bindent
        \FOR{$l=1$ to $L$}
        \STATE $\phi_l^{A[i]}  \sim   {\rm Dir}(\phi_l^{A[i]} \mid {\bf{O}}^A,{\bf{C}}^{A[i-1]},\beta^A)$
        \STATE $\phi_l^{B[i]} \sim {\rm Dir}(\phi_l^{B[i]} \mid {\bf{O}}^B,{\bf{C}}^{B[i-1]},\beta^B)$
        \ENDFOR
        \FOR{$k=1$ to $K$}
        \STATE $\theta_k^{A[i]} \sim {\rm Dir}(\theta_k^{A[i]} \mid {\bf{O}}^A,{\bf{W}}^{[i-1]},\alpha^A)$
        \STATE $\theta_k^{B[i]}\sim {\rm Dir}(\theta_k^{B[i]} \mid {\bf{O}}^B,{\bf{W}}^{[i-1]},\alpha^B)$
        \ENDFOR
        \FOR{$d=1$ to $D$}
        \STATE $c_d^{A[i]} \sim {\rm Multi}(c_d^{A[i]} \mid \theta_{w_d^{[i-1]}}^{A[i]}){\rm Multi}(o_d^A \mid \phi_{c_d^{A[i]}}^{A[i]})$
        \STATE $c_d^{B[i]} \sim {\rm Multi}(c_d^{B[i]} \mid \theta_{w_d^{[i-1]}}^{B[i]}){\rm Multi}(o_d^B \mid \phi_{c_d^{B[i]}}^{B[i]})$
        \STATE $w_d^{[i]} \sim {\rm Multi}(c_d^{A[i]} \mid \theta_{w_d^{[i]}}^{A[i]}){\rm Multi}(c_d^{B[i]} \mid \theta_{w_d^{[i]}}^{B[i]})$
        \ENDFOR
    \ENDFOR
    \end{algorithmic}
\end{algorithm}

\subsubsection{Computational model of the symbol emergence based on an inference procedure using the Metropolis-Hastings algorithm} 
A communication protocol based on the M-H algorithm proposed in SERKET enables us to develop a valid cognitive model, i.e., updating parameters by an agent does not require the agent to use cognitively unobservable information ~\citep{Serket,MH}. The M-H algorithm is one of Markov chain Monte Carlo algorithms, and Gibbs sampling is a special case of it. It is known that both algorithms can sample latent variables from the posterior distribution. That means that, theoretically, both of the algorithms can converge to the same stationary distribution. 

 Algorithm~\ref{alg2} shows the proposed inference algorithm based on the M-H algorithm.
 It can be also regarded as a semiotic communication between two agents, and individual object categorization process under the influence of words that are given by the other agent.
 
 A set of all the words $\bf{W}^{[i]}$ at $i_{th}$ iteration is calculated by two steps of the M-H algorithm, as shown in lines 3 and 5. Basically, the M-H algorithm requires information that an agent can observe within the dotted line in Figure~\ref{fig:model} and $w_d$.
In this model of symbol emergence, word $w_d$ is generated, i.e., uttered, by a speaker agent, either A or B.  A listener agent judges if the word properly represents the object the agent looks at. The criterion for the judgement should rely on the information the listener knows, i.e., the probability variables inside the dotted line in Figure~\ref{fig:model}.

  Algorithm~\ref{alg3} shows the M-H algorithm, where $Sp$ and $Li$ are the speaker and listener, respectively.
Generation of word $w_d$ from speaker's observation $o^{Sp}_d$ and category $c^{Sp}_d$, which the speaker regards as the target object, is modeled as a sampling process using $P(w^{Sp}_d|c_d^{Sp}, \theta _d^{sp})$. This sampling can be performed by using the information that is available to the speaker agent. 
In line 3, the sampling and judgment of words $W$ are performed with agent A as the speaker, and agent B as the listener. 
In line 5, the sampling and judgment of words $W$ are performed with agent B as the speaker, and agent A as the listener. 
In the M-H algorithm, the li that is available to the listener agents, i.e., $\cdot^{Li}$ and $w_d$. 

Simultaneous use of Algorithms 2 and 3 performs a probabilistic inference of the probabilistic generative models shown in Figure~\ref{fig:model}.
Importantly, the M-H algorithm can sample words from the posterior distribution exactly the same way as Gibbs sampling that requires all information  owned by an individual agent in a distributed manner.
This gives us a mathematical support of the dynamics of symbol emergence. 
\begin{algorithm}
\caption{Proposed interactive learning process based on the M-H algorithm}
\label{alg2}
    \begin{algorithmic}[1]
      \STATE Initialize all parameters
      \FOR{$i = 1$ to $I$}
        \STATE ${\bf W}^{[i]},{\bf{C}}^{A[i]},{\bf{C}}^{B[i]},\theta^{B[i]}$ = M-H algorithm(${\bf{O}}^{A},{\bf W}^{A[i-1]},{\bf{C}}^{A[i-1]},{\bf{O}}^{B},{\bf W}^{B[i-1]},{\bf{C}}^{B[i-1]},\theta^{B[i-1]}$)
        \STATE ${\bf W}^{B[i]} \leftarrow {\bf W}^{[i]}$
        \STATE ${\bf W}^{[i]},{\bf{C}}^{B[i]},{\bf{C}}^{A[i]},\theta^{A[i]}$ = M-H algorithm(${\bf{O}}^{B},{\bf W}^{B[i]},{\bf{C}}^{B[i]},{\bf{O}}^{A},{\bf W}^{A[i-1]},{\bf{C}}^{A[i]},\theta^{A[i-1]}$)
        \STATE ${\bf W}^{A[i]} \leftarrow {\bf W}^{[i]}$
      \ENDFOR
    \end{algorithmic}
\end{algorithm}
\begin{algorithm}
\caption{M-H algorithm}
\label{alg3}
    \begin{algorithmic}[1]
        %\Procedure
        \STATE {M-H algorithm}{(${\bf{O}}^{Sp},{\bf W}^{Sp},{\bf{C}}^{Sp},{\bf{O}}^{Li},{\bf W}^{Li},{\bf{C}}^{Li},\theta^{Li}$):}
    %\bindent
        \FOR{$l=1$ to $L$}
        \STATE $\phi_l^{Sp}  \sim   {\rm Dir}(\phi_l^{Sp} \mid {\bf{O}}^{Sp},{\bf{C}}^{Sp},\beta^{Sp})$
        \ENDFOR
        \FOR{$k=1$ to $K$}
        \STATE $\theta_k^{Sp} \sim {\rm Dir}(\theta_k^{Sp} \mid {\bf{O}}^{Sp},{\bf{W}}^{Sp},\alpha^{Sp})$
        \ENDFOR
        \FOR{$d=1$ to $D$}
            \STATE $c_d^{Sp} \sim {\rm Multi}(c_d^{Sp} \mid \theta_k^{Sp}){\rm Multi}(o_d^{Sp} \mid \phi_l^{Sp})$
        \ENDFOR
        \FOR{$d=1$ to $D$}
            \STATE $w_d^{Sp} \sim P(w_d^{Sp} \mid c_d^{Sp},\theta_k^{Sp})$
            \STATE $z \sim {\rm min} \left( 1,\frac{P(c_d^{Li} \mid \theta_k^{Li},w_d^{Sp})}{P(c_d^{Li} \mid \theta_k^{Li},w_d^{Li})} \right)$
            \STATE $u \sim {\rm Unif}(0,1)$
            \IF{$u \le z$}
                \STATE $w_d = w_d^{Sp}$
            \ELSE
                \STATE $w_d = w_d^{Li}$
            \ENDIF
        \ENDFOR
        \FOR{$l=1$ to $L$}
        \STATE $\phi_l^{Li} \sim {\rm Dir}(\phi_l^{Li} \mid {\bf{O}}^{Li},{\bf{C}}^{Li},\beta^{Li})$
        \ENDFOR
        \FOR{$k=1$ to $K$}
        \STATE $\theta_k^{Li} \sim {\rm Dir}(\theta_k^{Li} \mid {\bf{O}}^{Li},{\bf{W}},\alpha^{Li})$
        \ENDFOR
        \FOR{$d=1$ to $D$}
            \STATE $c_d^{Li} \sim {\rm Multi}(c_d^{Li} \mid \theta_k^{Li}){\rm Multi}(o_d^{Li} \mid \phi_l^{Li})$
        \ENDFOR
        \RETURN ${\bf W},{\bf{C}}^{Sp},{\bf{C}}^{Li},\theta^{Li}$
    \end{algorithmic}
\end{algorithm}

\subsubsection{Dynamics of symbol emergence and category formation} 
\begin{figure}[bt!]
	\begin{center}
	\includegraphics[width=\linewidth]{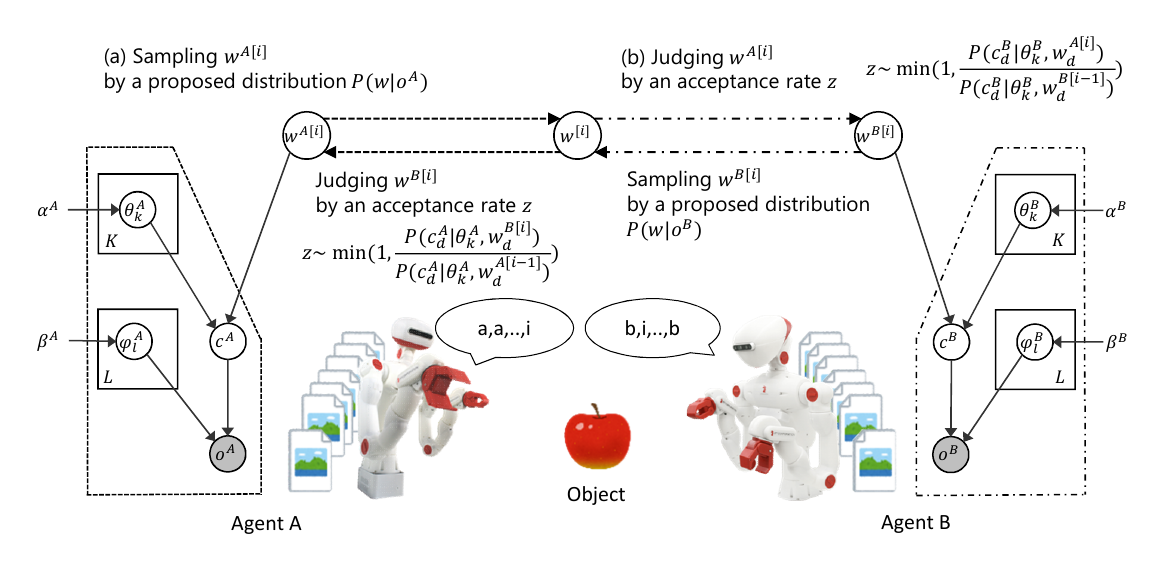}
	\caption{Dynamics of symbol emergence and category formation through semiotic communication between the agents in the proposed method.}
	\label{fig:overview}
	\end{center}
\end{figure}

This subsection describes how the proposed inference algorithm explains the dynamics of symbol emergence and category formation through semiotic communication.
Figure~\ref{fig:overview} conceptually shows the relationship between the dynamics of symbol emergence and concept formation between the agents, and the inference process for a word in the proposed model.
The proposed model consists of the categorization part, where the agents form categories individually, and the communication part, in which the agents exchange words between them. The categorization part is modeled based on latent Dirichlet allocation (LDA). The communication part connects the categorization parts of agents A and B. We modeled the communication part as the inference process of hidden variable $w_d^{[i]}$ in the M-H algorithm.

In line 12 of Algorithm~\ref{alg3}, word $w_d^{Sp}$ is sampled by a proposal distribution with parameters of the speaker only (i.e., $c_d^{Sp}$ and $\theta_k^{Sp}$) by the following formula:
 \begin{eqnarray}
 \label{w}
    w_d^{Sp} \sim P(w_d^{Sp} \mid c_d^{Sp},\theta_k^{Sp}).
 \end{eqnarray}

The process can be regarded as a word utterance from agent A in observation $o_d^{A}$ based on its internal parameters, as shown in Figure~\ref{fig:overview} (a). This is a part of semiotic communication.

In line 13, sampled word $w_d^{Sp}$ is judged by the listener by using acceptance rate $z$ calculated by the following formula:
  \begin{eqnarray}
  \label{z}
    z & \sim & {\rm min} \left( 1,\frac{P(c_d^{Li} \mid \theta_k^{Li},w_d^{Sp})}{P(c_d^{Li} \mid \theta_k^{Li},w_d^{Li})} \right).
  \end{eqnarray}

Acceptance rate $z$ of sampled word $w_d^{Sp}$ can be calculated from parameters of the listener only (i.e., $c_d^{Li},\theta_k^{Li},w_d^{Li}$). Therefore, this is plausible from a cognitive perspective.

In lines 14 to 19, sampled word $w_d^{Sp}$ is probabilistically accepted or rejected by the listener using acceptance rate $z$ and uniform random number $u$ by the following formulas:
  \begin{eqnarray}
    u \sim {\rm Unif}(0,1),
  \end{eqnarray}
  \begin{eqnarray}
      w_d^{[i]} = \begin{cases} 
      w_d^{Sp[i]} & (u \le z) \\
      w_d^{Li[i-1]} & (otherwise),
      \end{cases}
  \end{eqnarray}
where the continuous uniform distribution is denoted as ${\rm Unif}(\cdot )$. Word $w_d^{Lp[i-1]}$ of the listener at a previous iteration is used when sampled word $w_d^{Sp}$ is rejected.
Roughly speaking, if the listener agent considers that the current word is likely to be the word that represents the object that the listener also looks at, the listener agent accepts the word and updates its internal representations with a high probability.
The process can be explained as a judgment as to whether agent B accepts or rejects an utterance of agent A based on self-knowledge, as shown in Figure~\ref{fig:overview} (b).

In lines 21 to 29, the internal parameters of the listener are updated based on judged words $W$ by the following formulas:
  \begin{eqnarray}
    \phi_l^{Li} & \sim & {\rm Dir}(\phi_l^{Li} \mid {\bf{O}}^{Li},{\bf{C}}^{Li},\beta^{Li}),
  \end{eqnarray}
  \begin{eqnarray}
    \theta_k^{Li} & \sim & {\rm Dir}(\theta_k^{Li} \mid {\bf{O}}^{Li},{\bf{W}},\alpha^{Li}),
  \end{eqnarray}
  \begin{eqnarray}
    c_d^{Li} & \sim & {\rm Multi}(c_d^{Li} \mid \theta_{w_d}^{Li}){\rm Multi}(o_d^{Li} \mid \phi_{c_d^{Li}}^{Li}).
  \end{eqnarray}

The process can be explained as the updation of self-knowledge based on partial acceptance of the other agent's utterance. Because both the utterance and acceptance of words use only self-knowledge, these processes can be rationally convinced. 

As shown in Algorithm~\ref{alg2}, words $W$, categories $C$ and parameters $\phi_l$ and $\theta_k$ are inferred by repeating this process with $I$ iterations while exchanging the agents A and B.
This inference process not only is rationalized as a model of the symbol emergence and category formation through semiotic communication between the agents, but also gives a mathematical guarantee on the inference of the model parameters.

\section{Experiment}
\subsection{Experimental setup}
\subsubsection{Procedure}
\begin{figure}[bt!]
	\begin{center}
	\includegraphics[width=\linewidth]{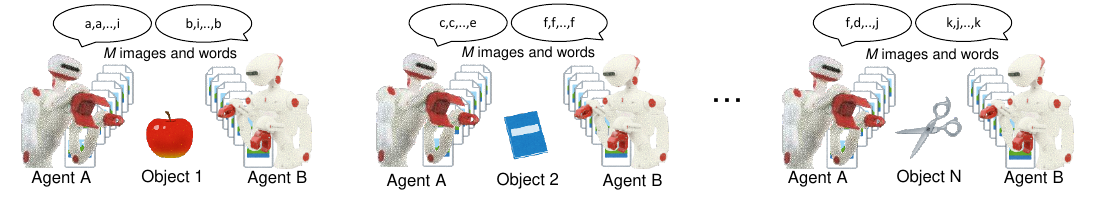}
	\caption{Overview of the experiment: agents A and B observed $N$ objects placed at the front of them. An agent captures images and suggests words $M$ times for each object.}
	\label{fig:experiment}
	\end{center}
\end{figure}

%実験目的・手順の説明
We performed an experiment to verify the validity of the proposed model and algorithm for modeling the dynamics of symbol emergence and concept formation. Specifically, we used an experiment of object categorization in the real world.
We also discuss the functions required for semiotic communication in the category formation and the symbol emergence in multi-agent systems from the comparison of three communication algorithms on the proposed model.
Figure~\ref{fig:experiment} shows an overview of the experiment.

The experiment was performed by the following procedure:
\begin{itemize}
  \item Step 1: Capture and memorize $N$ objects with $M$ images for each object on agent A and B with different angles.
  \item Step 2: Convert a memorized image to a visual feature as observations $o_d^A, o_d^B$ for agent A and B.
  \item Step 3: Sampling $w_d$ and updating model parameters from observations $o_d^A, o_d^B$ by the M-H algorithm. This step corresponds to semiotic communication between agents A and B based on the opponent's utterances and self-organized categories.
  \item Step 4: Repeat step 3 with $I$ iterations.
  \item Step 5: Evaluate the coincidence of words and categories between agents A and B for each object.
\end{itemize}
In the experiment, objects $N$, images $M$, and iterations $I$ were set as 10, 10, and 300, respectively.
We performed steps 1 to 5 with 10 trials for a statistical evaluation.

\subsubsection{Capturing and memorizing images (Step 1)}
\begin{figure}[bt!]
	\begin{center}
	\includegraphics[width=\linewidth]{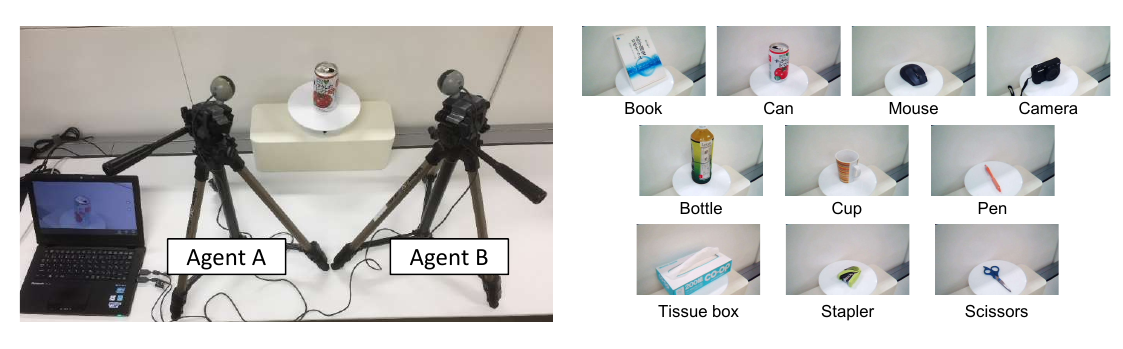}
	\caption{Experimental environment: two cameras were set as agent A and B with different angles. Each agent captured a target object placed at the center of two agents ten times, and we prepared ten objects as targets.}
	\label{fig:condition}
	\end{center}
\end{figure}

Figure~\ref{fig:condition} shows the experimental environment.
Two cameras on agents A and B captured object's images from different angles.
Captured images were memorized on a computer.
Resolution of a captured image was 640 pixels on width and 360 pixels on height.
Target objects were a book, can, mouse, camera, bottle, cup, pen, tissue box, stapler, and scissors , as shown in the right side of Figure~\ref{fig:condition}.

\subsubsection{Converting memorized images to observations (Step 2)}
An object's image captured by a camera is converted to a visual feature as an observation by Caffe~\citep{caffe}, which is a framework for convolutional neural networks (CNN)~\citep{CNN} provided by Berkeley Vision and Learning Center.
The parameters of CNN were trained by using the dataset from the ImageNet Large Scale Visual Recognition Challenge 2012\footnote{ILSVRC2012: http://www.image-net.org/challenges/LSVRC/2012/}.
Visual feature $o_i \in \{o_1,o_2,\cdots,o_{I}\}$ was calculated by the following equation:
\begin{equation}
\label{eq:vision}
{o}_i = \frac{\exp{(v_i)}}{\sum_{j=0}^I \exp{(v_j)}} \times 10^2,
\end{equation}
where $v_i \in \{v_1,v_2,\cdots,v_{I}\}$ is a value in the $7_{th}$ layer of Caffe. $I$ is the number of output units at the $7_{th}$ layer and was set as $4096$ in the experiment. We chose the $7_{th}$ layer to use Caffe as a feature extractor, not as an object recognizer. 

\subsubsection{Communication and categorization (Steps 3 and 4)}
We conducted the experiment by employing the proposed model with three communication algorithms:
\begin{itemize}
  \item No communication: $z=0$ is used instead of Formula~(\ref{z}) in the inference. In this algorithm, two agents have no communication. Each agent does the categorization based on observations only, without word information from the other agent.
  \item No rejection: $z=1$ is used instead of Formula~(\ref{z}) in the inference. In this algorithm, two agents accept all words from the other agent and update model parameters.
  \item Proposed algorithm: Formula~(\ref{z}) by the M-H algorithm is used in the inference. In this algorithm, two agents decide whether to accept or reject a word from the other agent based on self-organized categories.
\end{itemize}
The validity of the proposed model in the dynamics of symbol emergence and category formation and the functions required for semiotic communication are discussed by comparing the results between the proposed algorithm and two baseline algorithms, i.e. no rejection and no communication.

The hyperparameters of the proposed model were set as follows: $\alpha^A = 0.01$, $\alpha^B = 0.001$, $\beta^A = 0.01$, $\beta^B = 0.001$. The number of data points $D$ was 100. The number of categories and words were set as follows: $L = 15$, $K = 15$, to cover ten target objects. As characters corresponding to the word index, we used the following 15 characters: a, b, c, d, e, f, g, h, i, j, k, l, m, n, and o.

\subsubsection{Evaluation criteria (Step 5)}
We evaluated the performance of the proposed model of the symbol emergence and concept formation  using the following metrics: the kappa coefficient~\citep{Kappa} of words and adjusted rand index (ARI)~\citep{ARI} of categories between the agents.

The kappa coefficient was used as an evaluation criteria indicating the coincidence of words between agents A and B.
Kappa coefficient $\kappa$ was calculated by the following equation:
\begin{eqnarray}
    \kappa = \frac{C_o-C_e}{1-C_e},
\end{eqnarray}
where $C_o$ is the coincidence rate of words between agents, and $C_e$ is the coincidence rate of words between agents by random chance. The kappa coefficient is judged as follows: $Excellent:(\kappa > 0.8), Good:(\kappa > 0.6), Moderate:(\kappa > 0.4), Poor:(\kappa > 0.0)$.

The ARI was used as an evaluation criteria indicating the coincidence of categories between agents A and B.
The ARI was calculated by the following equation:
\begin{eqnarray}
    {\rm ARI} = \frac{\rm Index-Expected~Index}{\rm Max~Index-Expected~Index}.
\end{eqnarray}

Welch's t-test was used for statistical hypothesis testing between the proposed algorithm and two baseline algorithms, i.e. no communication and no rejection.

\subsection{Experimental results}
\begin{table}[tb!]
  \begin{center}
  \caption{Kappa coefficient on words and ARI on categories between agents A and B: the result is described with mean, standard deviation (SD), p-value, and t-test for 3 algorithms: no communication, no rejection, and the proposed algorithm. In the t-test, **: $(p<0.01)$, *: $(p<0.05)$, n.s.: not significant.}
    \begin{tabular}{c c c c c c c c c} \hline
     & \multicolumn{4}{c}{Kappa coefficient} & \multicolumn{4}{c}{ARI}\\
    Algorithm & Mean & SD & P-value & T-test & Mean & SD & P-value & T-test\\ \hline
    No communication & 0.01 & 0.04 & $8.9\times10^{-18}$ & ** & 0.89 & 0.07 & $2.2\times10^{-1}$ & n.s.\\
    No rejection & 0.57 & 0.06 & $1.0\times10^{-9}$ & ** & 0.88 & 0.03 & $3.0\times10^{-3}$ & **\\
    Proposed algorithm & \underline{\bf 0.88} & 0.05 & - & - & \underline{\bf 0.92} & 0.02 & - & -\\ \hline
    \end{tabular}
  \label{tab:coincidence}
  \end{center}
\end{table}
Table~\ref{tab:coincidence} shows the experimental results: the kappa coefficient and ARI for the proposed algorithm and two baseline algorithms, i.e. no communication and no rejection. 

For the kappa coefficient on words between agents A and B, the proposed algorithm obtained a higher value than the baseline algorithms, and there were significant differences between the proposed algorithm and baseline algorithms in the t-test. The result implies that the agents used the same words for observations with a very high coincidence (of 0.8 or more) in the proposed algorithm.

The ARI for the proposed algorithm was higher than for the baseline algorithms, and there were significant differences between the proposed algorithm and no rejection. In case of no rejection, the word has a negative effect on categorization between the agents, comparing with the result of no communication. On the other hand, in the proposed algorithm that stochastically accepts the other agent's word based on self-knowledge, the word positively acts on the categorization between agents. This result suggests that a rejection strategy in the semiotic communication works as an important function in the language evolution. Naturally, our result suggests it is biologically feasible and mathematically feasible.
\begin{figure}[bt!]
  \begin{tabular}{cc}
    \begin{minipage}[t]{0.47\hsize}
        \centering
        \includegraphics[scale=0.51]{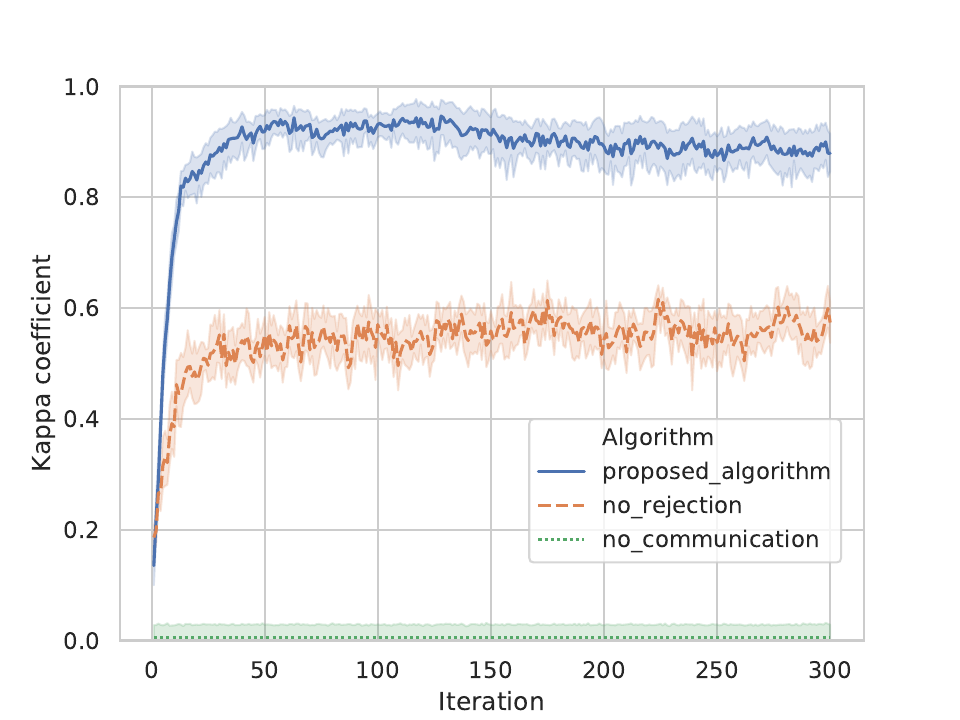}
        \caption{Transition on the kappa coefficient of words between agents: a line shows an average value, and top and bottom of each color show a maximum and minimum values in ten trials.}
        \label{fig:Kappa}
      \end{minipage} &
      \begin{minipage}[t]{0.47\hsize}
        \centering
        \includegraphics[scale=0.51]{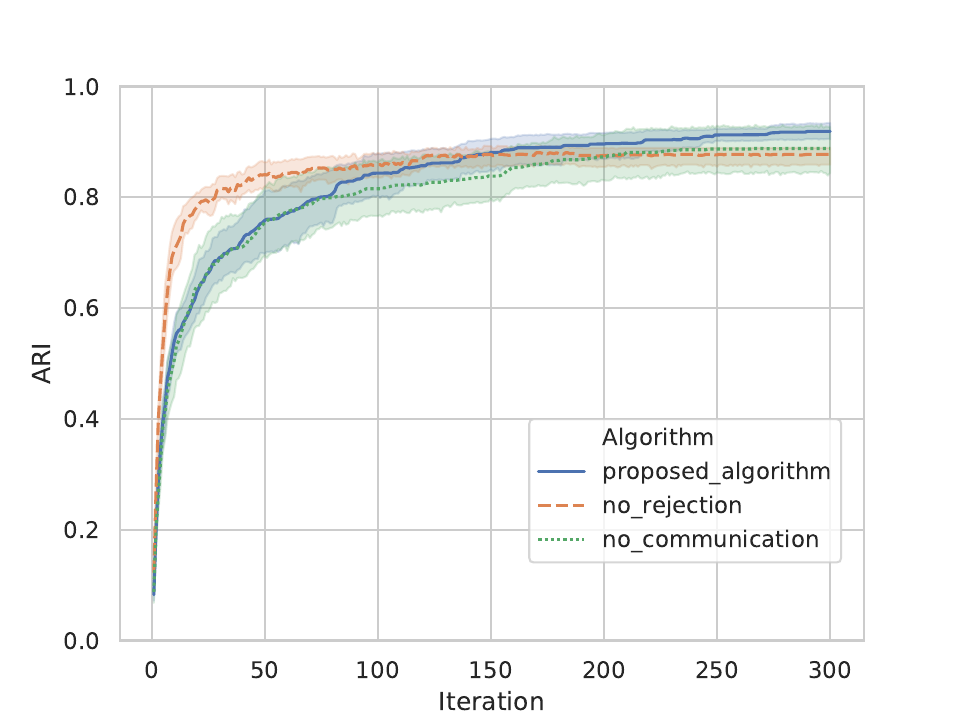}
        \caption{Transition on the ARI of categories between agents: a line shows an average value, and top and bottom of each color show a maximum and minimum values in ten trials.}
        \label{fig:ARI}
    \end{minipage}
  \end{tabular}
\end{figure}

Figure~\ref{fig:Kappa} shows transitions in the kappa coefficients of words between agents A and B by the proposed algorithm, no rejection, and no communication in ten trials.
In case of no communication, because all words of the opponent were rejected, the coincidence of words was at the level of random chance.
In case of no rejection, although the coincidence of words is increasing in the initial iterations, it drifts and stagnates at approximately $0.55$, which is a moderate value.
In case of the proposed algorithm, the kappa coefficient is higher than for the baseline algorithms, and the sharing of words accompanying the increase in iterations was confirmed.

Figure~\ref{fig:ARI} shows transitions for the ARIs of categories between agents A and B by the proposed algorithm, no rejection, and no communication in ten trials.
Compared with the baseline algorithms, it was confirmed that the proposed algorithm gradually and accurately forms and shares the categories between the agents.
Compared with no communication, the proposed algorithm promoted the sharing of categories from approximately 80 iterations, and obtained the highest ARI after approximately 150 iterations.
The result suggests the dynamics of symbol emergence and concept formation, where a symbol communication slowly affects the category formation in an agent and promotes sharing of the categories between the  agents.
It is a cognitively natural result: repetition of semiotic communication in the same environment gradually causes the sharing of categories between the agents.
\begin{figure}[bt!]
	\begin{center}
	\includegraphics[width=\linewidth]{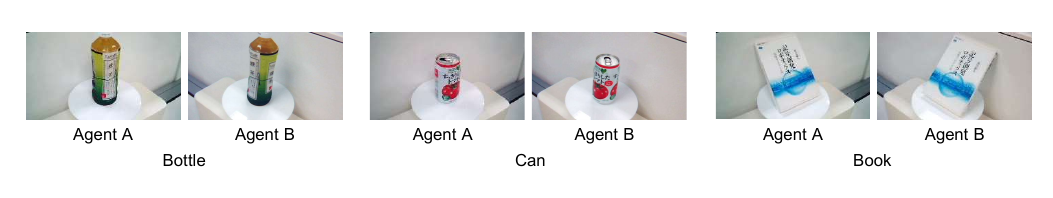}
	\caption{Examples of object's images: captured images of a bottle, can, and book from the viewpoints of agents A and B.}
	\label{fig:examples}
	\end{center}
\end{figure}

\begin{table}[tb!]
  \begin{center}
  \caption{Sampling results of words for three example objects by 3 communication algorithms, i.e. no communication, no rejection, and the proposed algorithm: the sampling results are described as $1_{st}$, $2_{nd}$, and $3_{rd}$ words in ten sampled words for each object. The rate of a word to ten sampled words is described in $(\cdot)$}
    \begin{tabular}{c c c c c c c c} \hline
     & & \multicolumn{2}{c}{No communication} & \multicolumn{2}{c}{No rejection} & \multicolumn{2}{c}{Proposed algorithm}\\
    Object & Word & Agent A & Agent B & Agent A & Agent B & Agent A & Agent B\\ \hline
     & $1_{st}$ & $h, l$ (0.2) & $j, l$ (0.2) & $b$ (0.7) & $i$ (0.5) & \underline{\bf $c$ (1.0)} & \underline{\bf $c$ (1.0)}\\
    Bottle & $2_{nd}$ & - & - & $i$ (0.2) & $b$ (0.4) & - & -\\
     & $3_{rd}$ & $b, d, f, i, m, o$ (0.1) & $b, d, e, h, i, k$ (0.1) & $c$ (0.1) & $c$ (0.1) & - & -\\ \hline
     & $1_{st}$ & $h, i$ (0.3) & $l$ (0.3) & $j$ (0.6) & $g$ (0.5) & \underline{\bf $f$ (1.0)} & \underline{\bf $f$ (1.0)}\\
    Can & $2_{nd}$ & - & $e$ (0.2) & $g, k$ (0.2) & $k$ (0.3) & - & -\\
     & $3_{rd}$ & $a, e, k, n$ (0.1) & $b, g, h, j, o$ (0.1) & - & j (0.2) & - & -\\ \hline
     & $1_{st}$ & $m, o$ (0.2) & $a$ (0.3) & $n$ (0.5) & $h$ (0.4) & \underline{\bf $b$ (0.6)} & \underline{\bf $b$ (0.6)}\\
    Book & $2_{nd}$ & - & $g, k$ (0.2) & $c$ (0.4) & $c, n$ (0.3) &  $k$ (0.4) &  $k$ (0.4)\\
     & $3_{rd}$ & $c, d, g, i, k, l$ (0.1) & - & $h$ (0.1) & - & - & -\\ \hline
    \end{tabular}
  \label{tab:word}
  \end{center}
\end{table}

For qualitative evaluation, we showed words assigned to each of the three objects: bottle, can, and book.
Figure~\ref{fig:examples} shows the examples of object's images observed from the viewpoints of agent A and B.
Table~\ref{tab:word} shows the words sampled by agents A and B for three example objects on three algorithms: the proposed algorithm and two baseline algorithms (no rejection and no communication).
The sampled words are described as three best results out of ten sampled words for each object. $(\cdot)$ shows the rate of a word to ten sampled words.
In case of no communication, a word representing an object was not shared between the agents.
In case of no rejection, words representing an object such as "i," "b," and "c" (for the bottle) were shared, but the probabilities of words are not high. 
In case of the proposed algorithm, it was confirmed that a word representing an object was shared between the agents with a high probability.

\begin{figure}[bt!]
 \begin{minipage}[b]{0.48\linewidth}
  \centering
  \includegraphics[scale=0.52]{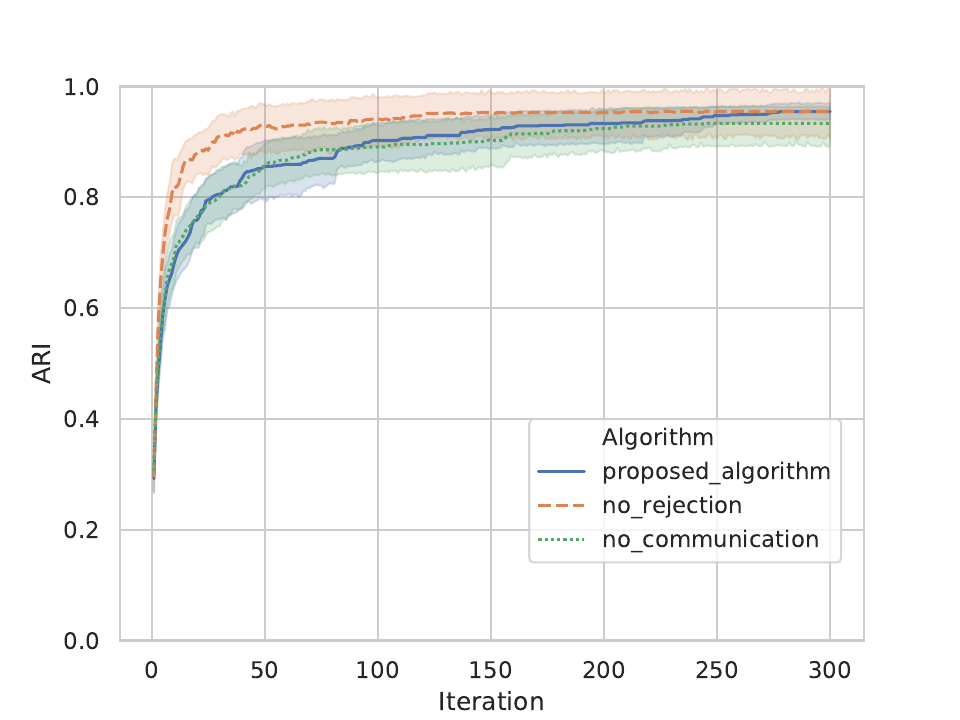}
  \subcaption{Agent A}
 \end{minipage}
 \begin{minipage}[b]{0.48\linewidth}
  \centering
  \includegraphics[scale=0.52]{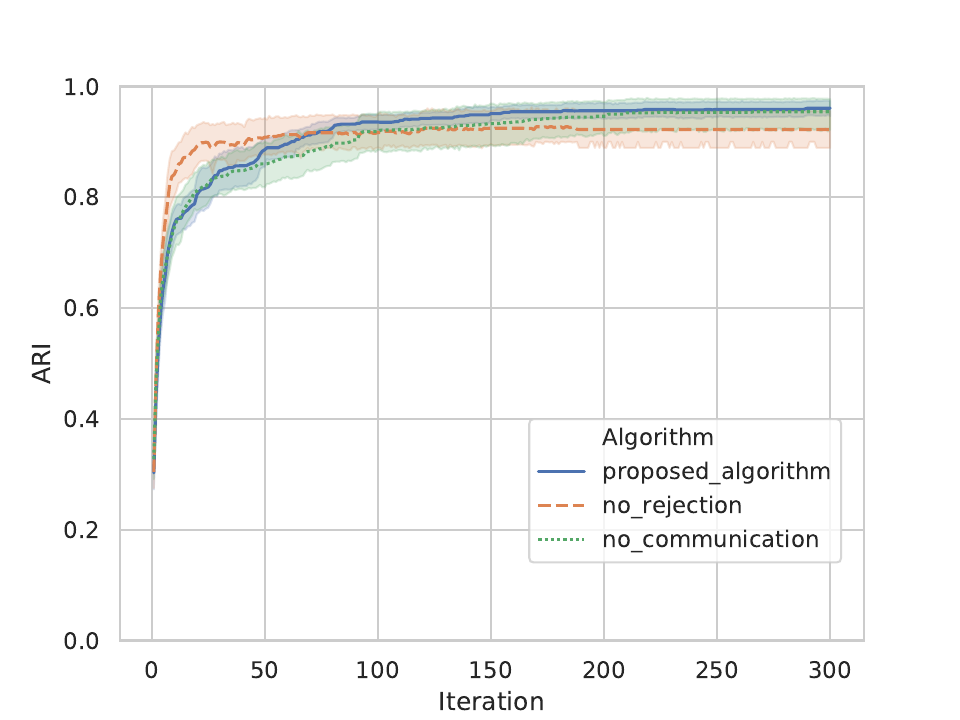}
  \subcaption{Agent B}
 \end{minipage}
 \caption{ARI between object labels and categories formed by the proposed algorithm in ten trials with agents A and B. The horizontal axis and vertical axis show the iteration and ARI, respectively.}
 \label{fig:examples_ARI}
\end{figure}

To evaluate the accuracy of categorization of actual objects, the ARI between the object labels and categories formed by the proposed algorithm is shown in Figure~\ref{fig:examples_ARI}.
At 300th iteration, the proposed algorithm shows a high accuracy, as the categorization of real-world objects by an unsupervised learning, despite being influenced by the words of another agent.

\begin{figure}[bt!]
 \begin{minipage}[b]{0.32\linewidth}
  \centering
  \includegraphics[scale=0.27]{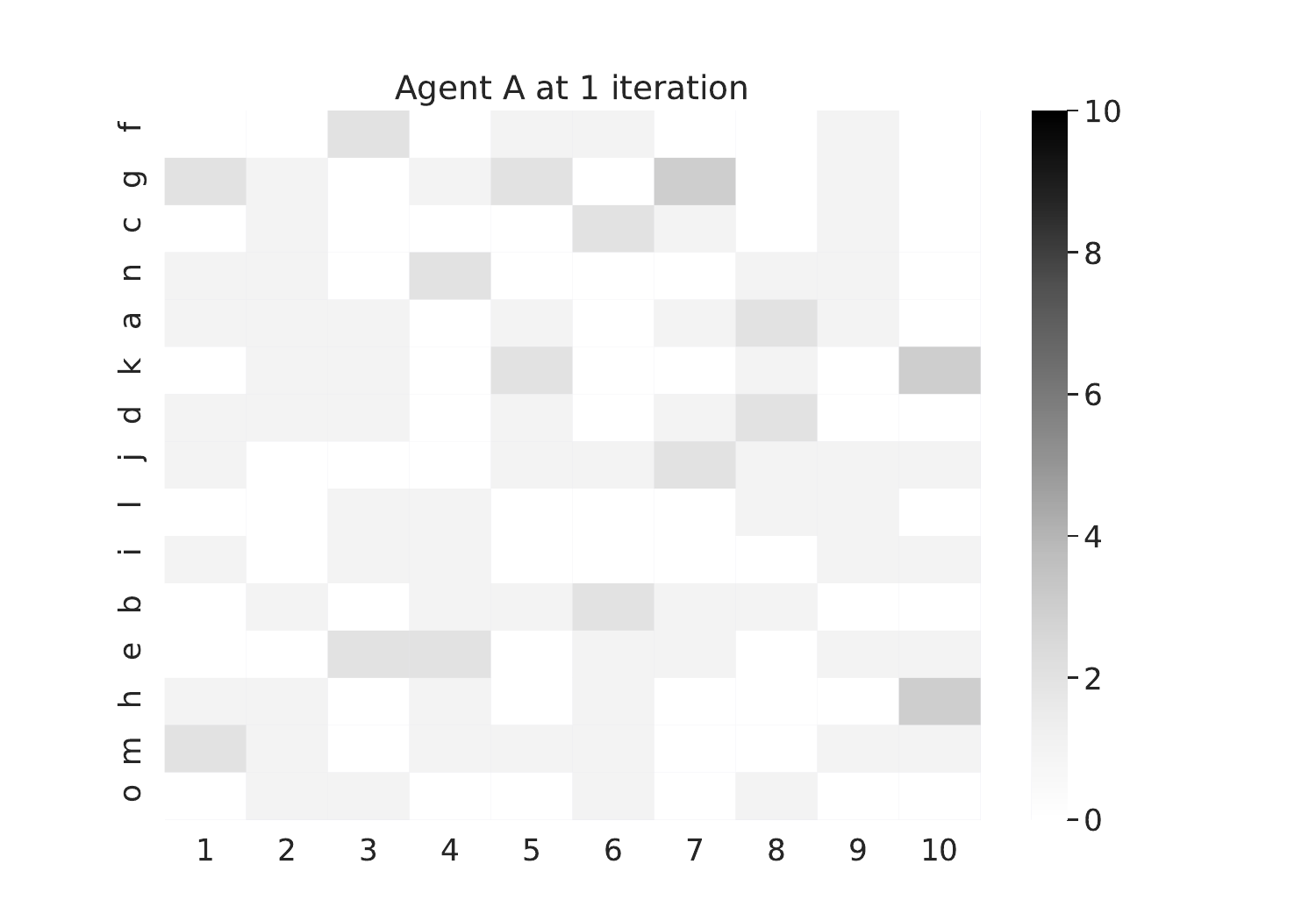}
  \subcaption{Agent A, 1 iteration}
 \end{minipage}
 \begin{minipage}[b]{0.32\linewidth}
  \centering
  \includegraphics[scale=0.27]{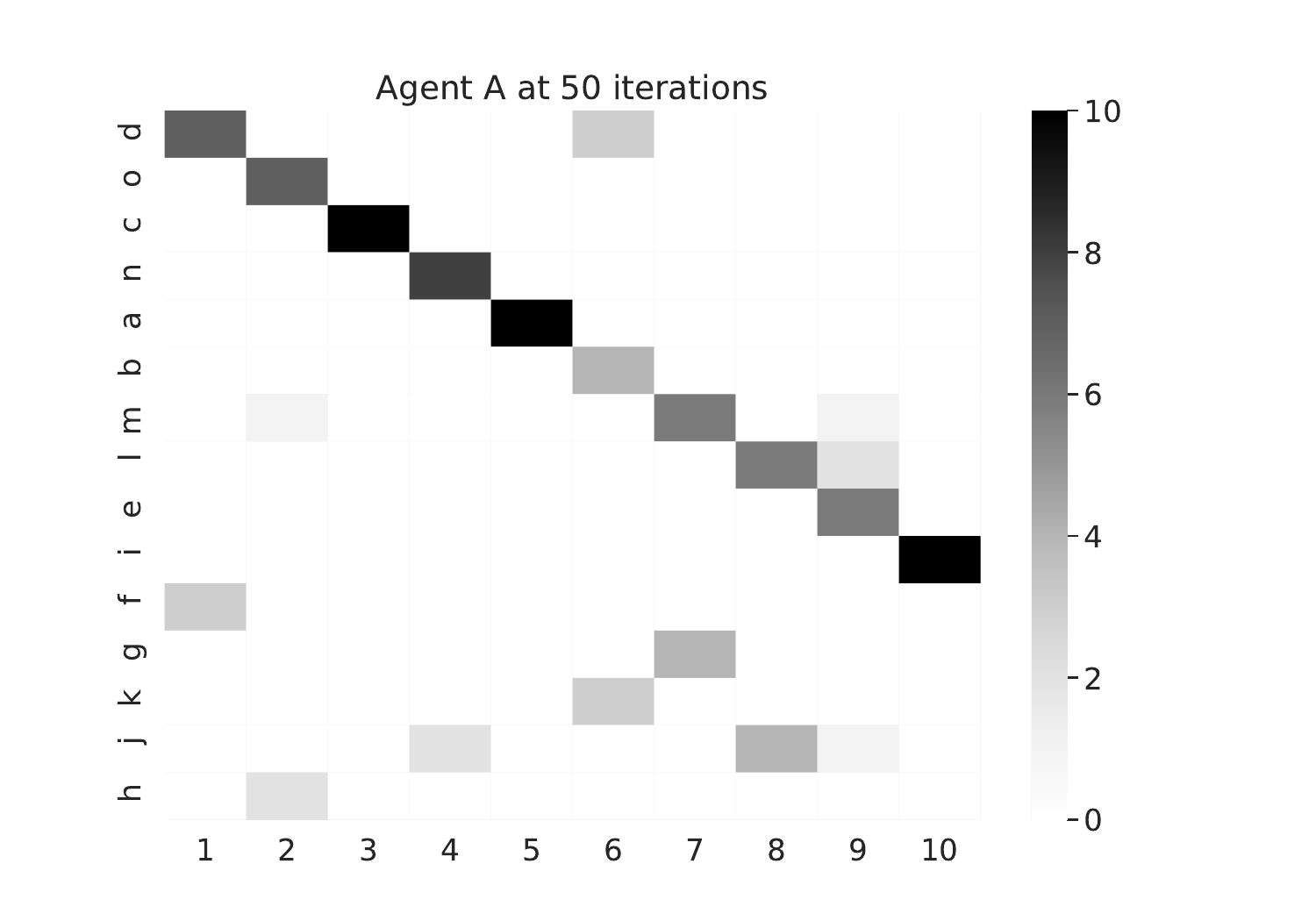}
  \subcaption{Agent A, 50 iterations}
 \end{minipage}
 \begin{minipage}[b]{0.32\linewidth}
  \centering
  \includegraphics[scale=0.27]{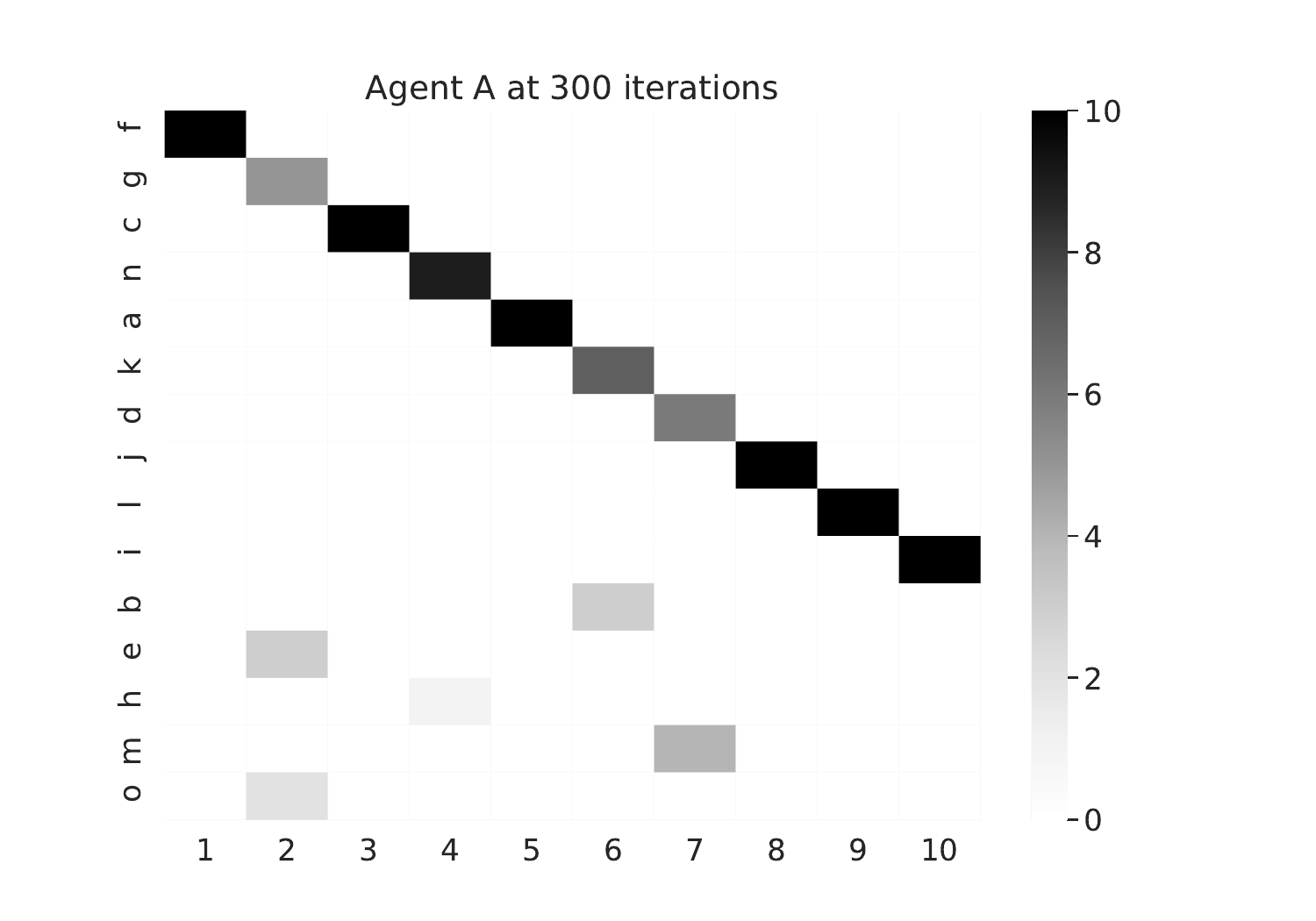}
  \subcaption{Agent A, 300 iterations}
 \end{minipage}\\
 \begin{minipage}[b]{0.32\linewidth}
  \centering
  \includegraphics[scale=0.27]{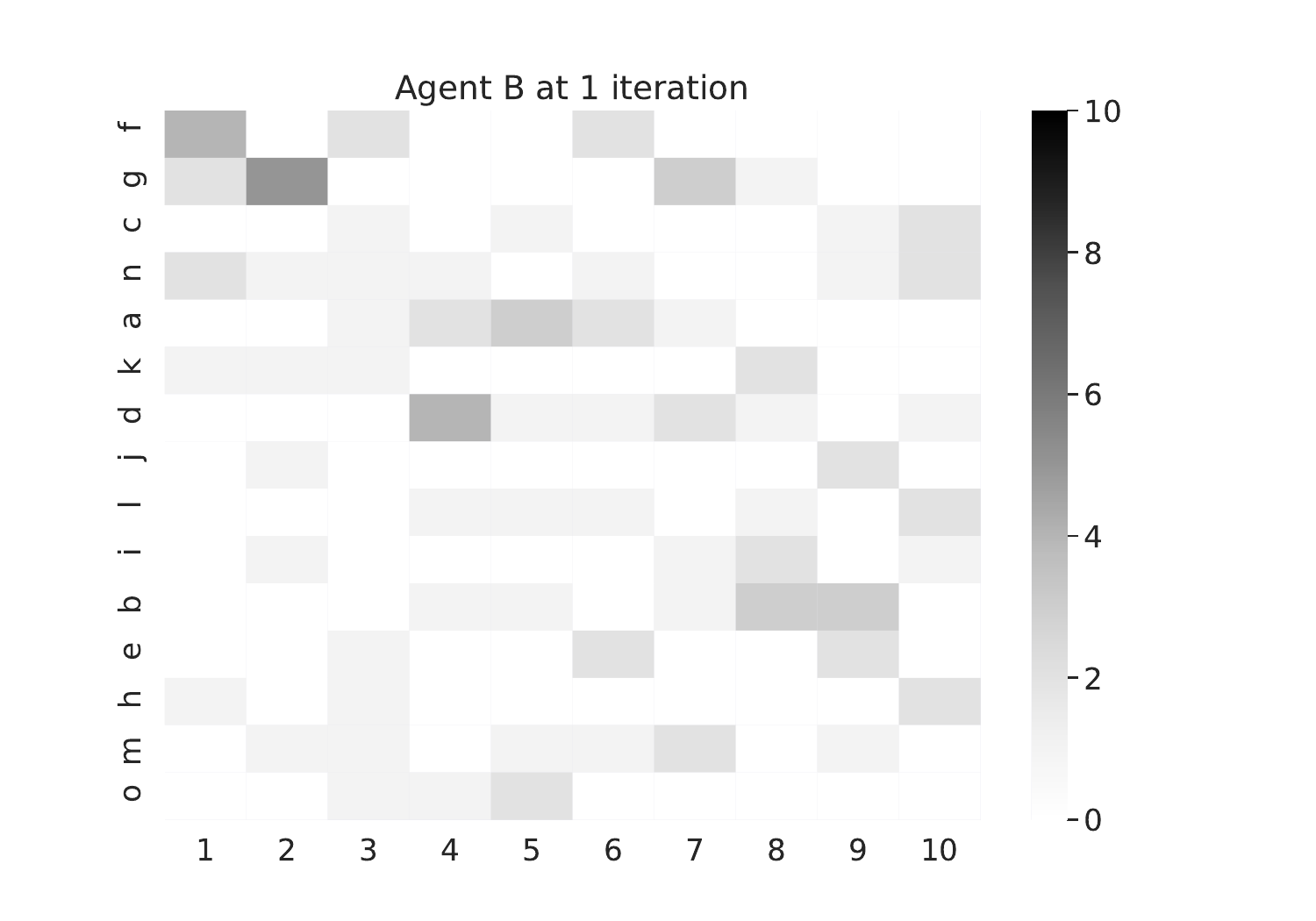}
  \subcaption{Agent B, 1 iteration}
 \end{minipage}
 \begin{minipage}[b]{0.32\linewidth}
  \centering
  \includegraphics[scale=0.27]{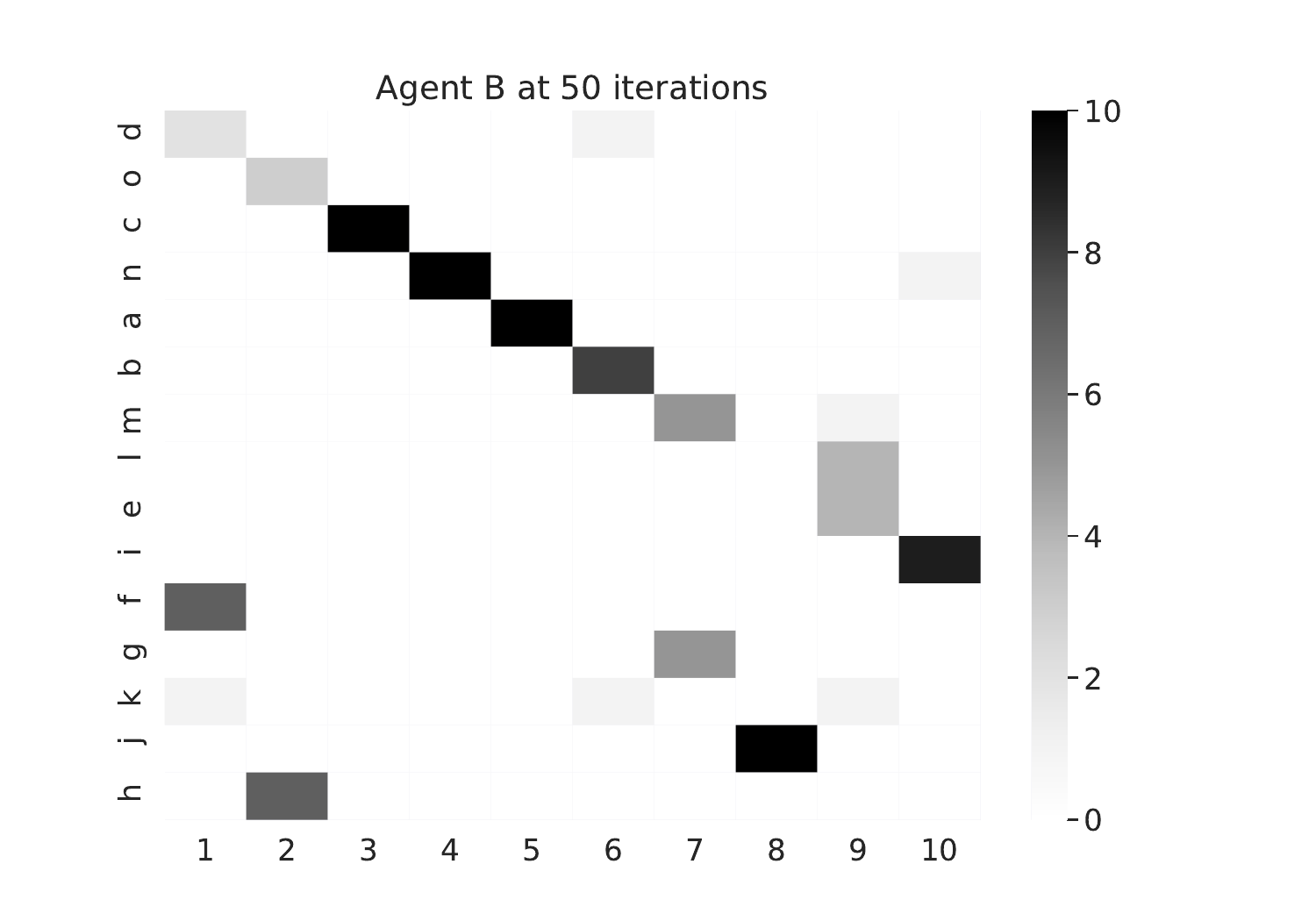}
  \subcaption{Agent B, 50 iterations}
 \end{minipage}
 \begin{minipage}[b]{0.32\linewidth}
  \centering
  \includegraphics[scale=0.27]{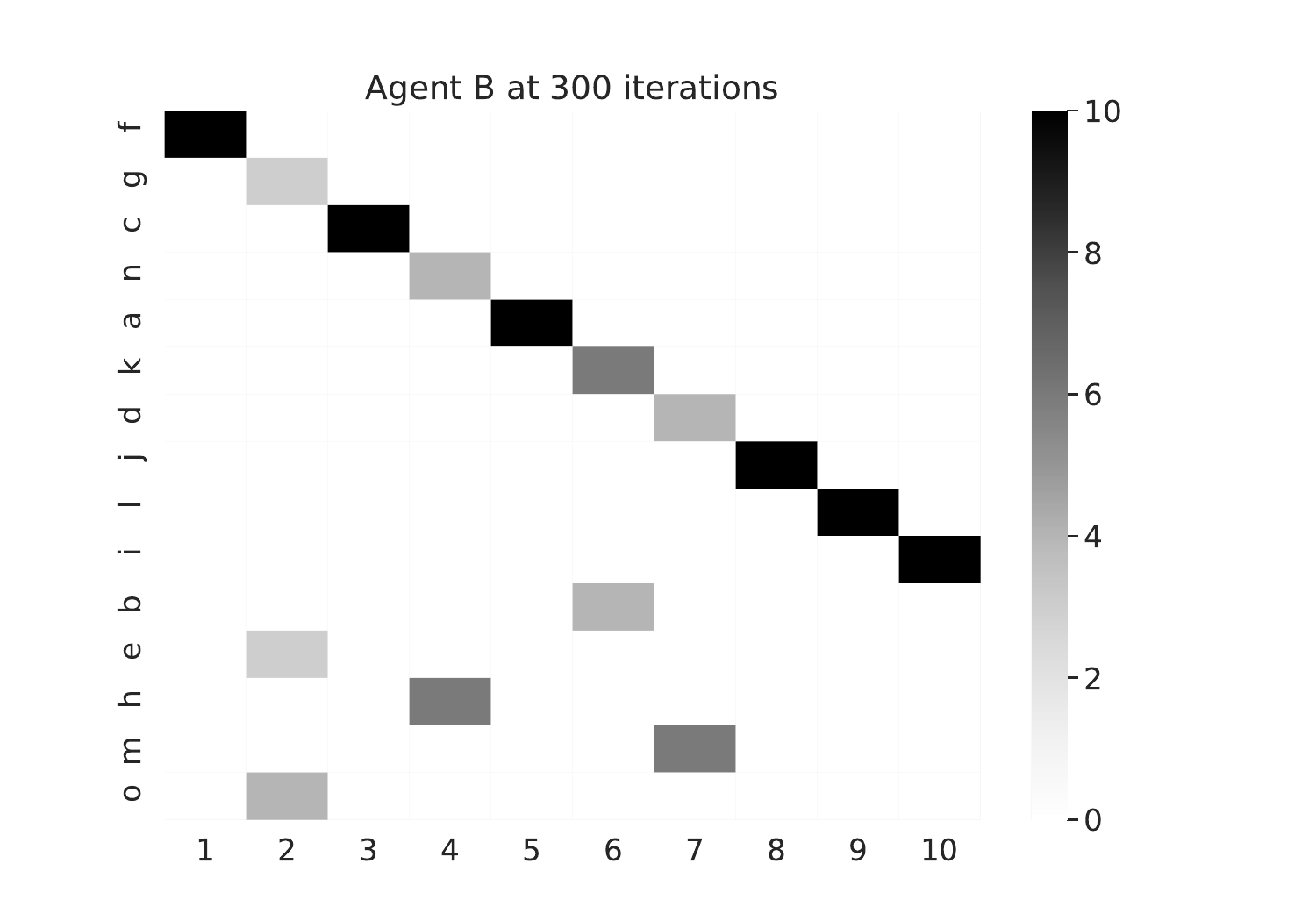}
  \subcaption{Agent B, 300 iterations}
 \end{minipage}
 \caption{Confusion  matrix between words and object's labels in each agent. The horizontal axis and vertical axis show the index of object's label and word, respectively. The order of the words was sorted according to the frequency of each object at agent A with 300 iterations.}
 \label{fig:matrix}
\end{figure}

The learning process of the correspondence relationship between words and objects for each agent is shown in Figure~\ref{fig:matrix} as a confusion matrix.
As the number of iterations increases, words corresponding to object labels were learned from random to one-on-one relationship. Each word was allocated to describe an object at the result of 300 iterations.

\begin{figure}[bt!]
 \begin{minipage}[b]{0.48\linewidth}
  \centering
  \includegraphics[scale=0.52]{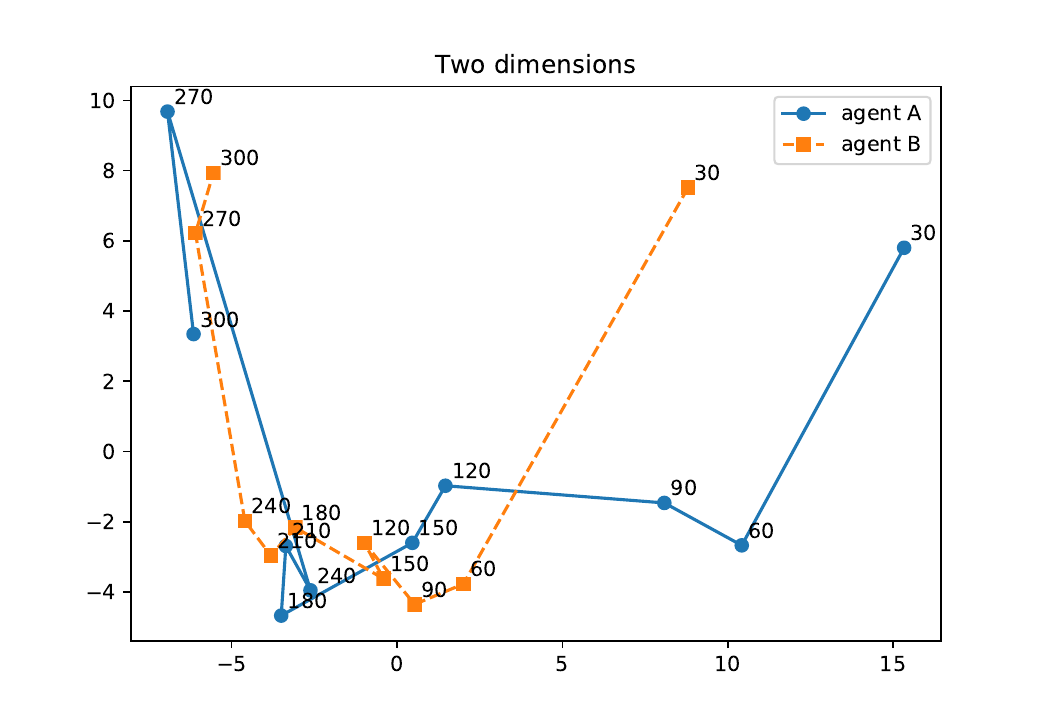}
  \subcaption{Two dimensions}
 \end{minipage}
 \begin{minipage}[b]{0.48\linewidth}
  \centering
  \includegraphics[scale=0.52]{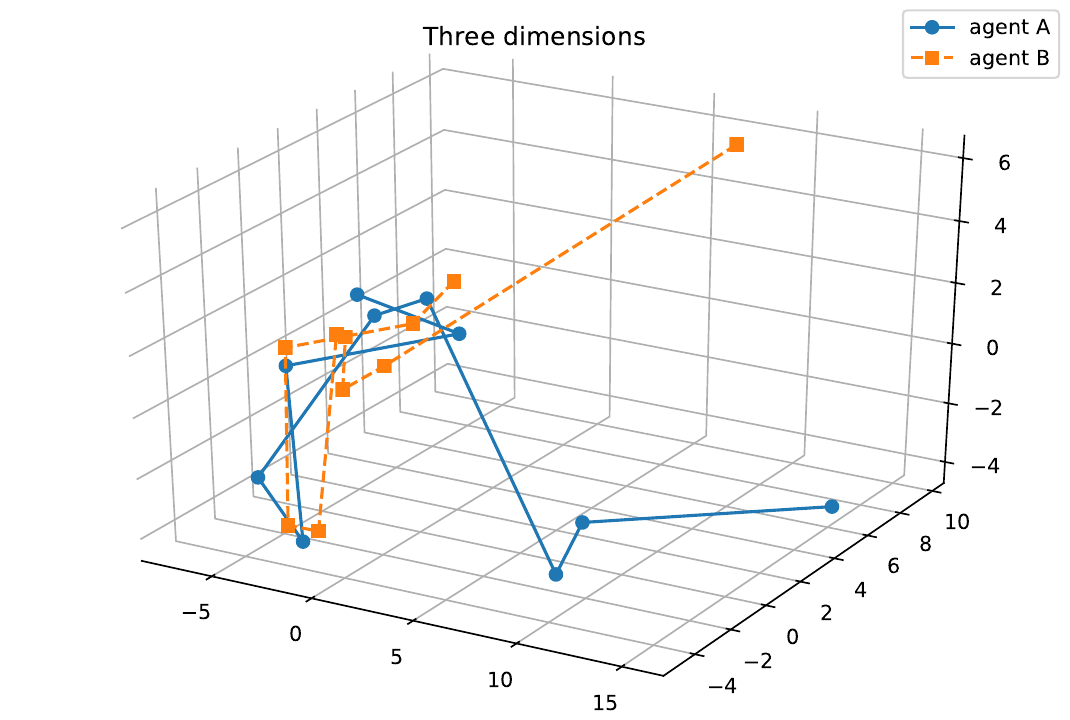}
  \subcaption{Three dimensions}
 \end{minipage}
 \caption{Results of PCA on the confusion matrix for agents A and B. The results are described from 30 to 300 iterations with a 30 iteration interval.}
 \label{fig:pca}
\end{figure}

Figure~\ref{fig:pca} shows the result of principal component analysis (PCA) between the confusion matrices for agents A and B in Figure~\ref{fig:matrix}. The results are described from 30 to 300 iterations at 30 iterations intervals on two and three dimensions. As the number of iterations increases, the results of PCA on the confusion matrices of two agents are getting closer. This can be interpreted as a process in which the interpretation system of words and objects between the agents approaches by the iteration of the semiotic communication.

\subsection{Discussion}
We evaluated the validity of the proposed model and algorithm as a model of the dynamics on symbol emergence and category formation from the experiments using daily objects in the real-world environment.
In the experiment, we compared the process of symbol emergence and category formation of objects between the agents by using three communication algorithms: the proposed algorithm, no rejection, and no communication.
The experimental results demonstrated the following three events in the communication algorithms.
\begin{itemize}
 \item In case of no communication, when the agent rejects all the other agent's utterances, the coincidence of categories was high but the coincidence of words was not shared between the agents. This result is understood as the following event: similar categories are formed when two agents have similar sensors that individually observe the same object.
 \item In case of no rejection, when the agent unconditionally accepts the other agent's utterances and updates the internal parameters, the coincidence of words drifts and stagnates, and the coincidence of categories decreases, compared with no communication. This result is understood as the following event: other agent's utterances that use different symbols interfere with categorization within the agent's individual as a noise.
  \item In the proposed algorithm, which probabilistically accepts the other agent's utterances based on the internal parameters, the coincidence of words was very high, and the coincidence of categories also had a high value compared with other algorithms. This result is also convincing as a mechanism of the symbol emergence and category formation based on the human semiotic communication.
\end{itemize}
Furthermore, it was suggested that the semiotic communication needs the function of rejecting other's utterances based on one's knowledge in the dynamics of symbol emergence and category formation between the agents. Naturally, our result can be interpreted as biologically and mathematically feasible.

\section{Conclusions}
This study focused on the symbol emergence in a multi-agent system and the category formation in individual agents through semiotic communication that is the generation and interpretation of symbols associated with categories formed from the agent's perception.
We proposed a model and an inference algorithm representing the dynamics of symbol emergence and category formation through semiotic communication between the agents as an interpersonal multimodal categorizer.
We showed the validity of the proposed model and inference algorithm on the dynamics of symbol emergence and concept formation in multi-agent system from the mathematical explanation and the experiment of object categorization and symbol emergence in a real environment.
The experimental results on object categorization using three communication styles, i.e. no communication, no rejection, and the proposed algorithm based on the proposed model suggested that semiotic communication needs a function of rejecting other's utterances based on one's knowledge in the dynamics of symbol emergence and category formation between agents.

This study did not model an emergence of a grammar. However, the proposed model and algorithm succeeded in giving a mathematical explanation for the dynamics of symbol emergence in multi-agent system and category formation in individual agents through semiotic communication.
This means our study showed a certain direction for treating multi-agent system logically in the symbol emergence and category formation.

As future work, we are extending the proposed model based on a mutual segmentation hypothesis of sound strings and situations based on co-creative communication~\citep{Okanoya07}.

The extension will be achieved through the following research process.
\begin{itemize}
 \item The extension for a mutual segmentation model of sound strings and situations based on multimodal information will be achieved based on a multimodal LDA with nested Pitman-Yor language model~\citep{mLDA} and a spatial concept acquisition model that integrates self-localization and unsupervised word discovery from spoken sentences~\citep{Akira16}.
 \item To reduce development and calculation costs associated with the large-scale model, ``Serket: An Architecture for Connecting Stochastic Models to Realize a Large-Scale Cognitive Model''~\citep{Serket}, will be used.
 \item Experiment with $N$ agents will be performed on symbol emergence and concept formation by expanding the proposed model. We can design an experiment as a communication structure based on human conversation, because human conversation is usually performed by two people. In a related study, \citet{Oshikawa18} proposed a Gaussian process hidden semi-Markov model, which enables robots to learn rules of interaction between persons by observing them in an unsupervised manner.
 \item Experimental results have shown the importance of a rejection strategy, but the evidence for the human brain to use such a strategy is not shown. We are planning to conduct psychological experiments.
 \item As an exploratory argument, mapping category $c$ to observation $o$  is theoretically possible for a neural network. A future study can develop a deep generative model, which integrates deep learning and generative model, by application of multimodal learning with deep generative models~\citep{Suzuki16}.
\end{itemize}

\section*{Funding}
This work was supported by MEXT/JSPS KAKENHI Grant Number JP17H06383 in \#4903 (Evolinguistics) and JP18K18134.

\bibliographystyle{frontiersinSCNS_ENG_HUMS} % for Science, Engineering and Humanities and Social Sciences articles, for Humanities and Social Sciences articles please include page numbers in the in-text citations
\bibliography{test}

\end{document}